\documentclass[sigconf, authorversion]{acmart}

\AtBeginDocument{%
  }

\copyrightyear{2026}
\acmYear{2026}
\setcopyright{cc}
\setcctype{by}
\acmConference[ETRA '26]{2026 Symposium on Eye Tracking Research and Applications}{June 01--04, 2026}{Marrakesh, Morocco}
\acmBooktitle{2026 Symposium on Eye Tracking Research and Applications (ETRA '26), June 01--04, 2026, Marrakesh, Morocco}
\acmDOI{10.1145/3797246.3806793}
\acmISBN{979-8-4007-2519-7/2026/06}

\acmSubmissionID{1011}

\usepackage{subfig}
\usepackage{listings}
\usepackage{algorithm}

\usepackage{booktabs}
\usepackage{longtable}

\usepackage{algorithmic}

\begin{document}

\title[Night Eyes]{Night Eyes: A Reproducible Framework for Constellation-Based Corneal Reflection Matching}

\author{Virmarie Maquiling}
\email{virmarie.maquiling@tum.de}
\affiliation{%
	\institution{Human-Centered Technologies for Learning, Technical University of Munich}
	\streetaddress{Marsstraße 20-22}
	\city{Munich} 
	\country{Germany} 
	\postcode{80335}
}
\affiliation{%
  \institution{Munich Center for Machine Learning (MCML)}
  \city{Munich}
  \country{Germany}
}

\author{Yasmeen Abdrabou}
\email{yasmeen.abdrabou@tum.de}
\affiliation{%
	\institution{Human-Centered Technologies for Learning, Technical University of Munich}
	\streetaddress{Marsstraße 20-22}
	\city{Munich} 
	\country{Germany} 
	\postcode{80335}
}
\affiliation{%
  \institution{Munich Center for Machine Learning (MCML)}
  \city{Munich}
  \country{Germany}
}

\author{Enkelejda Kasneci}
\email{enkelejda.kasneci@tum.de}
\affiliation{%
	\institution{Human-Centered Technologies for Learning, Technical University of Munich}
	\streetaddress{Marsstraße 20-22}
	\city{Munich} 
	\country{Germany} 
	\postcode{80335}
}
\affiliation{%
  \institution{Munich Center for Machine Learning (MCML)}
  \city{Munich}
  \country{Germany}
}

\renewcommand{\shortauthors}{Maquiling et al.}

\begin{abstract} 

Corneal reflection (glint) detection plays an important role in pupil-corneal reflection (P-CR) eye tracking, but in practice it is often handled as heuristics embedded within larger systems, making reproducibility difficult across hardware setups. We introduce a 2D geometry-driven, constellation-based pipeline for mulit-glint detection and matching, focusing on reproducibility and clear evaluation. Inspired by lost-in-space star identification, we treat glints as structured constellations rather than independent blobs. We propose a Similarity-Layout Alignment (SLA) procedure which adapts constellation matching to the specific constraints of multi-LED eye tracking. The framework brings together controlled over-detection, adaptive candidate fallback, appearance-aware scoring, and optional semantic layout priors while keeping detection and correspondence explicitly separated. Evaluated on a public multi-LED dataset, the system provides stable identity-preserving correspondence under noisy conditions. We release code, presets, and evaluation scripts to enable transparent replication, comparison, and dataset annotation.

\end{abstract}

\ccsdesc[500]{Human-centered computing~Interaction techniques}
\ccsdesc[500]{Computing methodologies~Matching}
\ccsdesc[300]{Computing methodologies~Image processing}

\begin{teaserfigure}
    \centering
  \includegraphics[width=.7\textwidth]{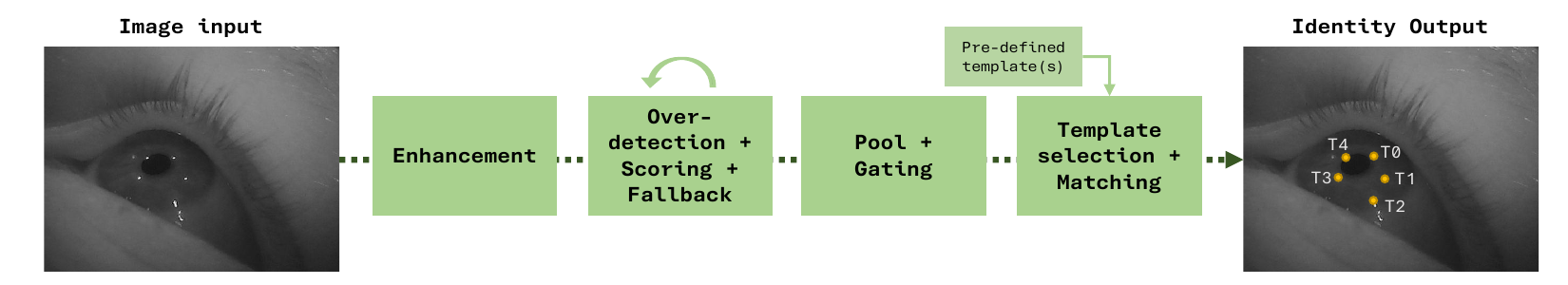} 
  \caption{Overview of the proposed multi-glint correspondence pipeline. The input image undergoes enhancement, over-detection with appearance scoring and adaptive fallback, candidate pooling and gating, and finally template-based matching using a predefined LED constellation. The output assigns identity-consistent glints to the detected reflections.} 
  \label{fig:teaser}
\end{teaserfigure}


\maketitle

\section{Introduction}
Pupil–corneal reflection (P–CR) eye tracking depends on accurate localization and matching of corneal reflections (glints)~\cite{guestrin2006general}. While pupil detection has been significantly improved in the past years, multi-glint detection remains challenging in practice. Glints can be unreliable, partly occluded, duplicated by spurious reflections, or disappear entirely. In head-mounted and virtual reality setups, multiple LEDS are arranged in fixed layouts~\cite{hua2007modeling, wang2015pupil}. Each glint needs to be matched to its corresponding light source to estimate gaze reliably. Many practical systems rely on straightforward techniques: thresholding for brightness, connected-component analysis, and geometric filtering relative to the pupil~\cite{mestre2018robust, zhao2016eye, geisler2018real}, and then applying hardware-specific rules about glint locations to decide which is which. However, the full matching pipeline is rarely described in detail in literature, and implementations are usually tailored to a particular hardware setup, making it difficult to compare approaches across devices or datasets. Deep learning approaches (e.g. \cite{chugh2021detection, maquiling2023v, byrne2025leyes, byrne2024precise}) have also been proposed, but they typically require carefully curated labeled or synthetic training data and often combine detection and correspondence within a single model. While efficient, this can make the system less interpretable and harder to adapt to systems with different LED layouts.

To address these issues, we introduce Night Eyes, a 2D geometry-driven, constellation-based framework for multi-glint detection and matching. Inspired by lost-in-space star tracking methods~\cite{rijlaarsdam2020survey, spratling2009survey}, we treat glints as a constellation rather than separate bright spots. While corneal reflections do not strictly satisfy the rigid assumptions used in star tracking, the matching problem ends up looking very similar: some reflections may be missing, extra ones may appear, and the correct configuration has to be recovered from noisy detections. Building on classical similarity alignment and geometric matching approaches used in constellation identification~\cite{mortari2004pyramid, umeyama2002least}, we introduce a Similarity-Layout Alignment (SLA) procedure designed specifically for multi-LED eye tracking. The pipeline combines controlled over-detection, adaptive candidate fallback, appearance-aware scoring, and a semantic layout prior into a single correspondence pipeline. The focus here is not on tuning a single configuration to squeeze out the highest possible score. Instead, the goal is to keep the system transparent and reproducible. Detection, scoring, and geometric matching are implemented as separate stages, which makes the behavior easier to interpret and evaluate across different illuminator layouts and recording conditions. To support reproducible experiments and future comparisons, we release the full codebase together with presets and evaluation scripts, as well as annotated glints for the OpenEDS datasets~\cite{garbin2019openeds, palmero2020openeds2020}. 

\section{Related Work}
Corneal reflections have traditionally been detected using relatively simple image processing techniques such as intensity thresholding and connected components~\cite{geisler2018real, mestre2018robust, zhao2016eye}. More recently, several works have explored deep neural networks for detecting glints and resolving their correspondence~\cite{chugh2021detection, maquiling2023v, byrne2025leyes, byrne2024precise}. Other approaches frame the problem as a graph matching task to improve robustness when detections are noisy or when spurious reflections are present~\cite{liu2022glints}. A useful parallel comes from star identification algorithms developed for space craft attitude estimation~\cite{rijlaarsdam2020survey, spratling2009survey, mortari2004pyramid}. These methods recover the identity of stars by matching geometric constellations, even when some stars are missing or false detections appear. Despite these advances, publicly available implementations with clearly defined evaluation procedures for identity-preserving glint correspondence remain relatively uncommon. For this reason, we design Night Eyes as a modular pipeline in which detection, scoring, and geometric correspondence are explicitly exposed as separable components. This makes it easier to compare different parts of the pipeline while evaluating the system under a fixed configuration and consistent metric definitions. 


\section{Pipeline Overview}

The pipeline operates per image and follows a staged design that separates enhancement, candidate generation, scoring, and 2D similarity-based geometric correspondence (see Figure~\ref{fig:teaser} for visualization). This separation is intentional: detection is allowed to over-generate plausible bright structures, while identity is handled later through geometric alignment. See Appendix~\ref{sec:alg1} for the full pipeline and SLA matcher algorithms.

\paragraph{Preprocessing and Enhancement.}
Each image is converted to grayscale and optionally denoised or contrast-adjusted. Small bright structures are amplified using either white top-hat filtering~\cite{huet1996textural}, Difference-of-Gaussians (DoG)~\cite{bundy1984difference}, or a high-pass operator, followed by optional CLAHE~\cite{pizer1987adaptive} and normalization. If enabled and a valid pupil center is available, a fixed-size pupil-centered ROI is extracted prior to detection. Otherwise, full-frame processing or frame skipping is applied according to the configured fail policy. The objective is not to detect exactly the number of ground truth glints, but to reliably amplify candidate reflections while suppressing background noise.

\paragraph{Candidate Extraction and Scoring.}
Candidate glints are detected using percentile thresholding on the enhanced image, followed by morphological cleanup and connected component extraction. For each component, simple geometric and intensity features are computed. These features are combined into a fixed-weight heuristic score that favors bright, compact, peak-like blobs. In \texttt{contrast\_support} mode, the score is augmented with normalized local contrast and Difference-of-Gaussians (DoG) responses, and further adjusted using geometric support voting among the top \texttt{support\_M} candidates in normalized coordinate space. This stage intentionally over-detects plausible blobs. At this stage, no geometric identity assumptions are applied yet.

\paragraph{Adaptive Fallback and Spatial Gating.}
If fewer than the target number of candidates are detected, additional passes are run with relaxed percentile thresholds and larger enhancement kernels. Candidates from all passes are merged within a fixed spatial radius, keeping the highest-scoring representative per cluster. After pooling the top-ranked candidates, optional border gating and pupil-centered annulus gating restrict implausible regions.

\paragraph{Template Construction.}
Templates are derived from labeled constellations using median aggregation or Procrustes alignment and normalized to zero mean and unit RMS scale. Matching can use either a single canonical template or a small template bank, with selection based on inlier count and residual consistency. Templates reflect the LED layout of the dataset and can be created by choosing a representative image containing all glints and manually annotating their locations. Matching parameters remain fixed afterward.

\paragraph{Constellation Matching via SLA}
Given the pooled candidates and a normalized template, correspondence is formulated as a 2D similarity alignment problem. Baseline matchers (RANSAC and star-based voting) are included for comparison. In the released implementation, matchers require at least three pooled candidates to run. Frames with fewer candidates return no match. The primary matcher, Similarity–Layout Alignment (SLA), follows constellation identification strategies used in star trackers. Starting from high-scoring candidates, SLA forms candidate triplets and checks whether their distance ratios are consistent with the template geometry. Valid triplets initialize similarity hypotheses that project the template into the image. The hypothesis then grows by assigning nearby candidates to template points within a pixel tolerance while repeatedly refitting the transform. Hypotheses that violate geometric consistency, scale limits, or residual thresholds are rejected. Optional layout and mirror constraints further penalize implausible configurations. The final match is selected based on inlier support and geometric consistency.

\paragraph{Evaluation.}
When labels are available, accuracy and precision are computed using a fixed pixel distance threshold (default = 10px). Localization error is measured for glints that are present and predicted, while identity-free metrics are computed by optimally matching predicted points to ground-truth points regardless of their assigned identities.

\section{Evaluation Protocol}
We evaluate correspondence in two ways: identity-preserving accuracy (correct LED-to-glint assignment) and identity-free accuracy (detection quality independent of labels), along with precision and pixel-level localization error. To configure the pipeline, we ran hyperparameter sweeps over the detection, scoring, and matching components using \citeauthor{chugh2021detection}'s public labeled mult-LED (5 glints) dataset~[\citeyear{chugh2021detection}], which includes ground truth pupil and glint locations. This dataset was used only for configuration selection and ablation analysis. All configurations were evaluated using the same tolerance thresholds and metric definitions, with diagnostic logging of common failure modes such as missed glints and ambiguous assignments. For reporting, we use a single configuration selected from the sweep based on a balance between identity accuracy, prediction, and residual error. This present uses the SLA matcher with semantic and mirror constraints enabled, along with adaptive fallback and geometric support during scoring. All results are reported using this frozen configuration.

\section{Results}
\begin{figure*}
    \centering
    \includegraphics[width=0.7\textwidth]{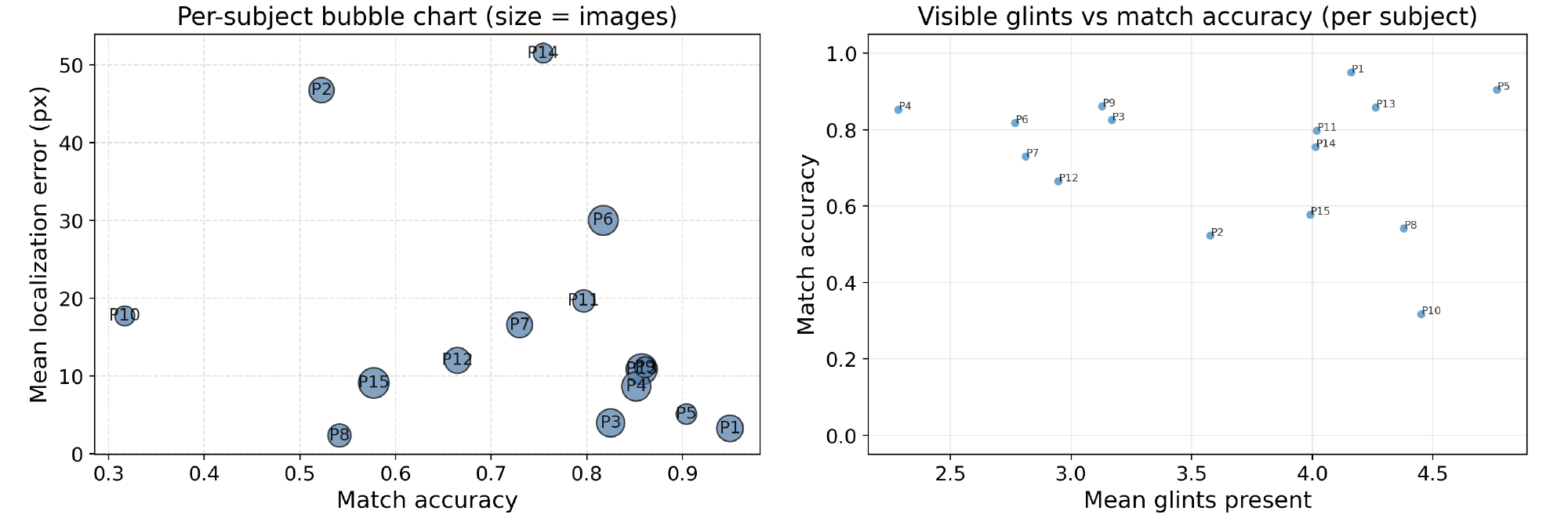}
    \caption{Per-subject results from the winning hyperparameter sweep evaluated on \citeauthor{chugh2021detection}'s dataset~\citeyear{chugh2021detection}}
    \label{fig:winsweep}
\end{figure*}

\begin{figure*}
    \centering
    \includegraphics[width=0.7\textwidth]{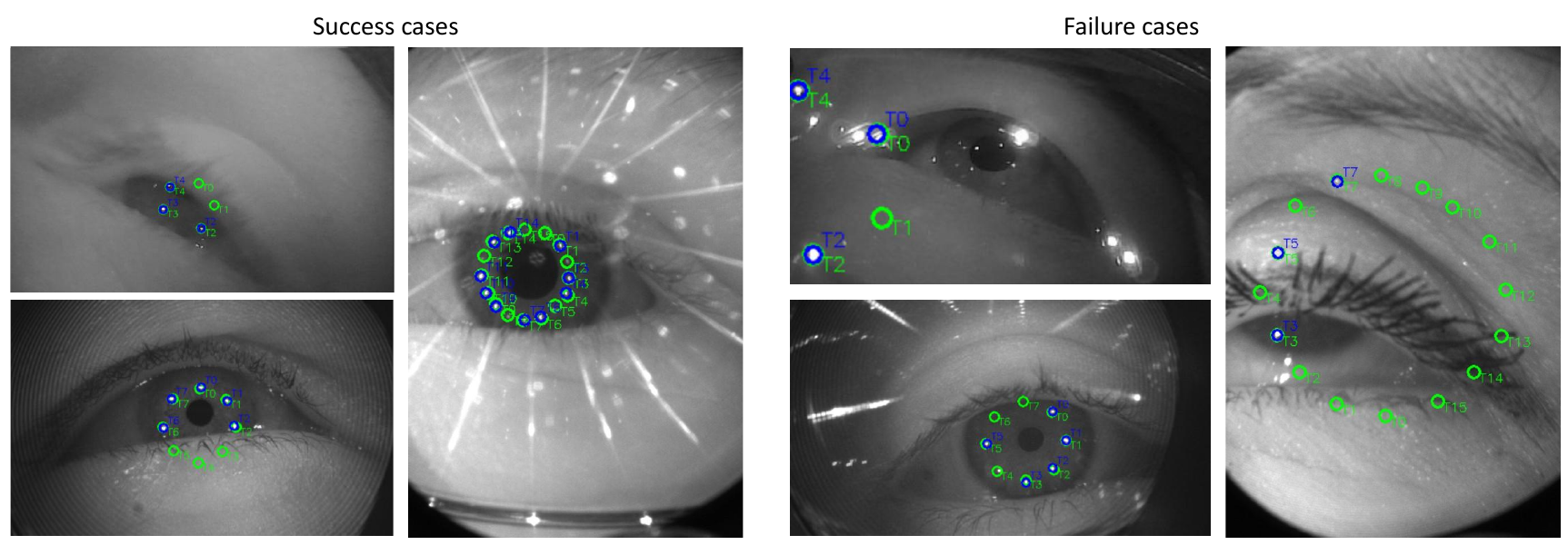}
    \caption{Success and failure cases across three eye trackers with different LED configurations. Images are from \citeauthor{chugh2021detection}'s dataset~\cite{chugh2021detection}, OpenEDS2019~\cite{garbin2019openeds}, and OpenEDS2020~\cite{palmero2020openeds2020}. Blue circles are the detected glints while the green circles show template-projected positions based on the chosen candidate glints.}
    \label{fig:cases}
\end{figure*}

We first report results on the labeled multi-LED dataset used during the hyperparameter sweeps. Using the frozen configuration, the system achieves 0.74 identity-preserving accuracy and 0.81 precision, with a median localization of 1.41 pixels. The mean error of 10.37 px is noticeably larger implying that when the system finds the correct correspondence, the predicted glints are very close to the ground truth, but a smaller number of failures produce large geometric errors that raise the average. The small gap between identity-preserving (0.74) and identity-free (0.77) accuracy suggests that most mistakes come from missed detections rather than incorrect LED label swaps. Performance also varies slightly across LEDs, likely due to occlusion (e.g. the eyelid covering the top and bottom glints) and viewing angle. Ablation results show that geometric support voting and adaptive candidate fallback help when some glints are missing, while the SLA semantic prior reduces mirrored or structurally implausible matches. A detailed breakdown of the sweeps and per-glint results for the selected configuration is provided in Appendix~\ref{app:sweep}. 

Figure \ref{fig:winsweep} (left) shows identity-preserving accuracy versus mean localization error per subject, with bubble size proportional to the number of images. Most subjects cluster in the high-accuracy, low-error region, although some variability appears under difficult illumination conditions or when several glints are missing. Interestingly, localization error stays fairly stable even when matching accuracy drops, suggesting that most failures arise during correspondence rather than detection.  Figure~\ref{fig:winsweep} (right) relates the average number of visible glints per subject to matching accuracy. While subjects with more visible glints generally show higher matching accuracy, the relationship isn't strictly monotonic. Some subjects with four or more reflections still exhibit lower accuracy, while others achieve reliable correspondence with only two or three glints. This shows that the geometric layout of the glints matter more than how many are detected in the image. 

To evaluate generalization, we apply the same frozen configuration to OpenEDS 2019~\cite{garbin2019openeds} and OpenEDS 2020 Sparse Segmentation~\cite{palmero2020openeds2020}. The only change is that the SLA layout prior is disabled, since it was tuned to the dataset of \citeauthor{chugh2021detection}~[\citeyear{chugh2021detection}]. Dataset-specific templates are constructed from labeled glint constellations. Because OpenEDS 2020 does not provide ground truth pupil location for all images, the pupil ROI step was disabled for that dataset. Despite differences in illumination and noise, the pipeline shows similar detection and matching behavior across datasets. This suggests that separating over-detection from geometric alignment transfers well across recording conditions. Some failures still occur, most commonly when predicted glints latch onto incorrect candidates and form a rotated constellation. Examples of success and failure cases across datasets are shown in Figure~\ref{fig:cases}.


\subsection{Discussion and Limitations}
The proposed framework is 2D geometry-driven and does not rely on learned representations. This makes the system easier to interpret and portable across different LED layouts, but it may limit performance in very low-contrast conditions where deep learning models may excel better in. The current implementation also assumes a fixed template structure and similarity-based alignment, which may be less suitable when reflections become highly distorted (e.g., in eye tracking setups with highly off-axis cameras, the constellation may warp considerably when it hits the sclera). Runtime grows with the number of SLA candidate seeds, although pooling limits keep computation manageable in practice. Using the default settings on \citeauthor{chugh2021detection}'s dataset, the pipeline runs at around 2.5 FPS ($\approx 0.39$ s/frame) on a CPU-only system, and can run faster using multiprocessing during batch processing. The SLA layout prior used in the frozen configuration was originally tuned with the Pupil Labs LED-configuration used by \citeauthor{chugh2021detection}~[\citeyear{chugh2021detection}]. For the OpenEDS experiments, this prior was disabled, yet the rest of the pipeline still performed consistently, suggesting that the detection--alignment design transfers across datasets even without dataset-specific layout assumptions. Another useful aspect of the pipeline is the projected template, which estimates where glints are expected to appear in the image once part of the constellation has been matched. This provides additional geometric cues even when some reflections are missing and could be useful for future algorithm development. Although the results on the OpenEDS are encouraging, testing the pipeline on a wider range of hardware setups would help further support its ability to generalize.

\section{Availability and Reproducibility}

The full implementation of the proposed pipeline is publicly available at \href{https://github.com/vbmaq/nighteyes}{https://github.com/vbmaq/nighteyes}. The repository includes the SLA matcher, candidate scoring modules, adaptive fallback mechanism, evaluation scripts, and UIs for template creation, full-pipeline execution, and annotation review. The UIs can be used off-the-shelf for researchers looking to create an annotated glint dataset for their own eye tracking prototypes. The codebase will be maintained and further developed to support reproducibility and future improvements. We also release the generated glint annotations (both detected glints and template-projected positions) for OpenEDS 2019~\cite{garbin2019openeds} and OpenEDS 2020 Sparse Segmentation~\cite{palmero2020openeds2020} at the following link: \href{https://doi.org/10.5281/zenodo.18847443}{https://doi.org/10.5281/zenodo.18847443} with minor corrections on correspondence errors. These extend OpenEDS beyond pupil, iris, and sclera segmentation by providing structured multi-glint correspondence labels aligned with the datasets' fixed LED layouts. All reported results can be reproduced using the frozen configuration and the documented command-line scripts. The system relies only on standard open-source dependencies. The annotations follow the CC BY-NC 4.0 license of the original OpenEDS datasets and are restricted to non-commercial use. They are derived from publicly released data and do not introduce additional personal information. Use remains subject to the original OpenEDS license and consent conditions. A detailed dataset card and machine-readable metadata are provided with the public release.

\section{Conclusion}
We presented Night Eyes, a reproducible 2D geometry-driven constellation matching framework for multi-glint detection and correspondence in P–CR eye tracking. Rather than relying purely on layout-specific heuristics or data-dependent models, we frame glint matching as a geometric alignment problem with explicit scoring, adaptive over-detection, and optional semantic constraints. Through systematic sweep analysis and a frozen configuration, we provide a transparent reference baseline evaluated on a labeled multi-LED dataset~\cite{chugh2021detection} and transferred to the OpenEDS datasets~\cite{garbin2019openeds, palmero2020openeds2020}. The released pipeline includes template construction, frozen configuration presets, and evaluation scripts. Although templates are derived from each dataset’s LED layout, the correspondence logic and hyperparameters remain unchanged, supporting reproducible cross-dataset evaluation. 

\section*{Acknowledgement}
The project is supported by the Chips Joint Undertaking (Chips JU) and its members, including top-up funding by Denmark, Germany, Netherlands, Sweden, under grant agreement No. 101139942.

\bibliographystyle{ACM-Reference-Format}
\bibliography{sample-base}

\newpage
\appendix
\section*{Appendix}

%


\section{Sweep Summaries and Configuration Selection}
\label{app:sweep}

Tables~\ref{tab:app_sweep_best} and \ref{tab:app_tierB2_collapsed} summarize the coarse (Tier~A) and refined (Tier~B) hyperparameter sweeps used to identify a stable operating configuration. Table~\ref{tab:app_sweep_best} reports the best-performing run per sweep stage. Table~\ref{tab:app_tierB2_collapsed} collapses the Tier~B2 sweep into unique performance outcomes and reports the associated fallback settings (FB, shown as percentile schedule / pass / target). All runs in Table~\ref{tab:app_tierB2_collapsed} use the SLA matcher with semantic prior and mirror rejection enabled; support-voting settings were fixed at Sup $M=20/30$, Sup tol $=0.08/0.10$, Sup $w=0.10/0.15$, with Ratio tol $=0.10$ and Pivot $P=6$. Table~\ref{tab:app_glint_breakdown} shows the per-glint performance of the frozen Tier~B2 configuration. Here, Acc.\ denotes identity-preserving recall, Prec.\ denotes precision, IDF denotes identity-free accuracy, and Med.Err.\ denotes median localization error in pixels.

\begin{table}[H]
\centering
\small
\caption{Best-performing configuration from each sweep stage.}
\setlength{\tabcolsep}{6pt}
\resizebox{\columnwidth}{!}{%
\begin{tabular}{ll l r r r r}
\toprule
Sweep stage & Best run ID & Matcher & Acc.\ $\uparrow$ & Prec.\ $\uparrow$ & ID-free Acc.\ $\uparrow$ & Med.\ Err.\ (px) $\downarrow$ \\
\midrule
Tier A (coarse) & \texttt{tierA/run\_008} & hybrid & 0.7289 & 0.8082 & 0.7787 & 1.416 \\
Tier B1 & \texttt{b1\_005} & sla & 0.7319 & 0.7956 & 0.7760 & 1.423 \\
Tier B2 (winner) & \texttt{tierB/b2\_081} & sla & 0.7414 & 0.8116 & 0.7740 & 1.414 \\
Tier B3 & \texttt{tierB/b3\_004} & sla & 0.7381 & 0.8042 & 0.7680 & 1.416 \\
\bottomrule
\end{tabular}%
}
\label{tab:app_sweep_best}
\end{table}

\begin{table}[H]
\caption{Collapsed Tier~B2 sweep summary with fallback settings.}
\centering
\scriptsize
\setlength{\tabcolsep}{3pt}
\renewcommand{\arraystretch}{1.05}
\begin{tabular}{l c r r r r l}
\toprule
Run ID & Ties & Acc. & Prec. & IDF & Med.Err. & FB \\
\midrule
\texttt{tierB/b2\_081} & 8 & 0.7414 & 0.8116 & 0.774 & 1.414 & 99,98,97 / 4 / 8 \\
\texttt{tierB/b2\_113} & 8 & 0.7364 & 0.8081 & 0.767 & 1.415 & 99.5,99,98.5,98 / 4 / 12 \\
\texttt{tierB/b2\_065} & 8 & 0.7346 & 0.8147 & 0.766 & 1.414 & 99,98,97 / 2 / 8 \\
\bottomrule
\end{tabular}
\label{tab:app_tierB2_collapsed}
\end{table}

\begin{table}[H]
\caption{Per-glint performance for the frozen Tier~B2 configuration.}
\centering
\small
\setlength{\tabcolsep}{6pt}
\renewcommand{\arraystretch}{1.1}
\resizebox{\columnwidth}{!}{%
\begin{tabular}{l r r r r r r r}
\toprule
Glint & Present & Pred. & Correct & Acc. $\uparrow$ & Prec. $\uparrow$ & Mean Err. (px) & Med. Err. (px) \\
\midrule
G0 & 1522 & 1485 & 1127 & 0.740 & 0.759 & 10.324 & 1.568 \\
G1 & 1845 & 1623 & 1382 & 0.749 & 0.852 & 8.060  & 1.350 \\
G2 & 1789 & 1576 & 1319 & 0.737 & 0.837 & 10.699 & 1.302 \\
G3 & 1737 & 1553 & 1256 & 0.723 & 0.809 & 11.003 & 1.407 \\
G4 & 1870 & 1768 & 1413 & 0.756 & 0.799 & 11.726 & 1.477 \\
\bottomrule
\end{tabular}%
}
\label{tab:app_glint_breakdown}
\end{table}

\section{Algorithms}
\label{sec:alg1}

\begingroup
\footnotesize

\refstepcounter{algorithm}
\noindent\rule{\columnwidth}{0.4pt}

\noindent\textbf{Algorithm~\thealgorithm} Per-image glint detection and matching pipeline
\label{alg:pipeline}

\noindent\rule{\columnwidth}{0.4pt}
\begin{algorithmic}[1]
\STATE \textbf{Input:} Image $I$, single template $T_0$ or bank $\mathcal{T}$, params $\theta$, optional labels $L$
\STATE \textbf{Output:} Predicted glints $\hat{G}$

\STATE $I_g\leftarrow\text{toGray}(I)$;\ $\theta'\leftarrow\text{scaleParams}(\theta,\text{size}(I_g))$
\STATE $roi\_active\leftarrow\text{false}$;\ $\Delta\leftarrow(0,0)$

\IF{pupil ROI enabled}
  \STATE $(c,r,ok)\leftarrow\text{detectPupilCenter}(I_g,\theta')$
  \STATE $(action,c_{\text{roi}})\leftarrow\text{resolvePupilROI}(c,r,ok,\theta')$ \COMMENT{may use last-good center (SP)}
  \IF{$action=\texttt{skip}$}
    \STATE \textbf{return} \texttt{None}
  \ENDIF
  \IF{$action=\texttt{use}$}
    \STATE $(I_g,\Delta)\leftarrow\text{computePupilROI}(I_g,c_{\text{roi}},\theta')$;\ $roi\_active\leftarrow\text{true}$
  \ENDIF
\ENDIF

\STATE $(C,w,raw)\leftarrow\text{detectCandidatesOnePass}(I_g,\theta')$ \COMMENT{enhance$\rightarrow$thresh$\rightarrow$CC$\rightarrow$score2}

\IF{fallback enabled \textbf{and} $raw<N_{\text{target}}$}
  \FOR{$i=1..P_{\max}$}
    \STATE $p\leftarrow\texttt{pcts}[\min(i{-}1,|\texttt{pcts}|{-}1)]$;\ $k\leftarrow i\cdot\texttt{cand\_fallback\_kernel\_add}$
    \STATE $(C_i,w_i,\_)\leftarrow\text{detectCandidatesOnePass}(I_g,\theta',p,k)$
    \STATE $(C,w)\leftarrow$
    \STATE \hspace{1.5em}$\text{mergeDedupKeepBest}(C\cup C_i,w\cup w_i,\texttt{cand\_merge\_eps})$
    \IF{$|C|\ge N_{\text{target}}$}
      \STATE \textbf{break}
    \ENDIF
  \ENDFOR
\ENDIF

\IF{$roi\_active$}
  \STATE $(C,w)\leftarrow\text{mapToFullAndInBounds}(C,w,\Delta)$
\ENDIF
\STATE $(C,w)\leftarrow\text{poolTopN}(C,w,N_{\max})$

\IF{border ROI gate}
  \STATE $(C,w)\leftarrow\text{filterCandidatesBorder}(C,w,\theta')$
\ENDIF
\IF{pupil annulus gate}
  \STATE $(c_g,r_g)\leftarrow\text{resolveGateCenterRadius}(c,r,ok,roi\_active,\theta')$
  \STATE $m\leftarrow\text{pupilAnnulusMask}(C,c_g,r_g,\theta')$;\ $s\leftarrow\sum m$
  \IF{$s\ge\texttt{min\_k}$ \textbf{or} \texttt{pupil\_force\_gate}}
    \STATE $(C,w)\leftarrow\text{applyMask}(C,w,m)$
  \ENDIF
\ENDIF

\STATE $R^\star \leftarrow$
\STATE \hspace{1.5em}$\begin{cases}
\arg\max_{T\in\mathcal{T}} \text{scoreMatchResult}(\text{runMatcher}(T,C,w,\theta'),\theta') & \text{(bank)}\\
\text{runMatcher}(T_0,C,w,\theta') & \text{(single)}
\end{cases}$

\STATE $\hat{G}\leftarrow\text{extractMatches}(R^\star)$
\IF{\texttt{post\_id\_resolve} \textbf{and} $|T_0|{=}5$ \textbf{and} $\text{fullFiveMatch}(R^\star)$}
  \STATE $\hat{G}\leftarrow\text{resolveIdentityPermutation}(\hat{G},\theta')$
\ENDIF
\IF{$L$ provided}
  \STATE compute metrics
\ENDIF
\STATE \textbf{return} $\hat{G}$
\end{algorithmic}
\noindent\rule{\columnwidth}{0.4pt}

\vspace{0.75\baselineskip}

\refstepcounter{algorithm}
\noindent\rule{\columnwidth}{0.4pt}

\noindent\textbf{Algorithm~\thealgorithm} Similarity--Layout Alignment (SLA) matcher
\label{alg:sla}

\noindent\rule{\columnwidth}{0.4pt}
\begin{algorithmic}[1]
\STATE \textbf{Input:} Template $T\in\mathbb{R}^{K\times2}$, candidates $C$, scores $w$, tolerance $\varepsilon$, params $\theta$
\STATE \textbf{Output:} Best match $R^\star$ or failure
\STATE \textbf{Note:} semantic/layout priors assume $K{=}5$.

\STATE $\mathcal{I}\leftarrow$ ratio index of $T$
\STATE $piv \leftarrow$ top \texttt{pivot\_P} candidates by $w$
\STATE compute effective ratio tolerance $\tau$ (adaptive + clamped if enabled)
\STATE $S\leftarrow\emptyset$

\FORALL{$c_p \in piv$}
  \STATE $S_p \leftarrow \emptyset$
  \FORALL{candidate pairs $(c_a,c_b)$}
    \STATE compute ratio $r$
    \FORALL{template triplets from ratio range-query within $\tau$}
      \STATE fit 3-point similarity $\hat{\mathcal{S}}$
      \IF{\texttt{sla\_semantic\_prior} and hard semantic/mirror veto triggers}
        \STATE \textbf{continue}
      \ENDIF
      \STATE add seed with quality weighted by appearance and seed residual
    \ENDFOR
  \ENDFOR
  \STATE keep top \texttt{max\_seeds\_per\_pivot} in $S_p$; $S \leftarrow S \cup S_p$
\ENDFOR
\STATE keep global top \texttt{max\_seeds} seeds in $S$

\FORALL{seed transform $\mathcal{S}$}
  \STATE initialize matches $M$ from seed
  \WHILE{new matches accepted}
    \STATE for each unmatched template point:
    \STATE \hspace{1.5em}try candidates within $\varepsilon$ (distance then $w$), refit
    \STATE \hspace{1.5em}accept if median residual $\le$ \texttt{grow\_resid\_max}
  \ENDWHILE

  \STATE perform final assignment (greedy/Hungarian), refit
  \STATE reject if median residual $>\varepsilon$, max residual $>2\varepsilon$, or scale gate fails

  \STATE compute $err=$ median residual; compute appearance sum and optional layout/semantic penalties
  \STATE select best by: max inliers $\rightarrow$ min cost $\rightarrow$ max appearance $\rightarrow$ min $err$
\ENDFOR

\IF{$|R^\star| < \texttt{min\_inliers}$}
  \STATE \textbf{return} failure
\ELSE
  \STATE \textbf{return} $R^\star$
\ENDIF
\end{algorithmic}
\noindent\rule{\columnwidth}{0.4pt}

\endgroup

\end{document}